\documentclass{article} %
\usepackage{preprint,times}

\usepackage{amsmath,amsfonts,bm}

\def\figref#1{figure~\ref{#1}}

\def\secref#1{section~\ref{#1}}

\def\eqref#1{equation~\ref{#1}}

\def\1{\bm{1}}

\DeclareMathAlphabet{\mathsfit}{\encodingdefault}{\sfdefault}{m}{sl}
\SetMathAlphabet{\mathsfit}{bold}{\encodingdefault}{\sfdefault}{bx}{n}

\usepackage{hyperref}
\usepackage{url}

\usepackage[ruled,vlined]{algorithm2e}
\usepackage{amsmath,amssymb}
\usepackage{graphicx}
\usepackage{xcolor}
\renewcommand{\paragraph}[1]{\vspace{-1.5mm}{\flushleft\textbf{#1}}}

\usepackage{caption}
\usepackage{graphicx}
\usepackage[table]{xcolor}
\usepackage{makecell}
\usepackage{amsmath}
\usepackage{tcolorbox}
\usepackage{multirow}
\usepackage{enumitem}
\usepackage{pifont}

\newcommand{\model}[0]{SpatialSpeak}

\definecolor{navyblue}{HTML}{0071BC}
\definecolor{hotpink}{HTML}{FF0080}
\definecolor{oai-white}{HTML}{FFFFFF}
\definecolor{oai-black}{HTML}{000000}
\definecolor{oai-red}{HTML}{FF4500}
\definecolor{oai-green}{HTML}{51DA4C}
\definecolor{oai-blue}{HTML}{0000FF}
\definecolor{oai-yellow}{HTML}{FFF639}
\definecolor{oai-magenta}{HTML}{FF45FF}
\definecolor{oai-cyan}{HTML}{00FFFF}
\definecolor{oai-orange}{HTML}{FE7600}
\definecolor{oai-violet}{HTML}{8A2BE2}
\definecolor{oai-brown}{HTML}{A0522D}
\definecolor{oai-green-050}{HTML}{F4FFF4}
\definecolor{oai-green-100}{HTML}{E9FFE8}
\definecolor{oai-green-200}{HTML}{D9FFD8}
\definecolor{oai-green-300}{HTML}{C9FFC7}
\definecolor{oai-green-400}{HTML}{A6FFA3}
\definecolor{oai-green-500}{HTML}{7CF178}
\definecolor{oai-green-600}{HTML}{51DA4C}
\definecolor{oai-green-700}{HTML}{3FA93B}
\definecolor{oai-green-800}{HTML}{2D712A}
\definecolor{oai-green-900}{HTML}{193718}
\definecolor{oai-gray-000}{HTML}{FFFFFF}
\definecolor{oai-gray-100}{HTML}{FAFAFA}
\definecolor{oai-gray-200}{HTML}{F5F5F5}
\definecolor{oai-gray-300}{HTML}{E5E5E5}
\definecolor{oai-gray-400}{HTML}{FFB7A4}
\definecolor{oai-gray-500}{HTML}{CDCDCD}
\definecolor{oai-gray-600}{HTML}{A8A8A8}
\definecolor{oai-gray-700}{HTML}{747474}
\definecolor{oai-gray-800}{HTML}{393939}
\definecolor{oai-gray-900}{HTML}{000000}

\definecolor{mygreen}{HTML}{3cb44b}

\definecolor{myred}{HTML}{E33222}
\definecolor{gr}{RGB}{0, 146, 0}

\graphicspath{{images/}}

\usepackage{makecell}
\usepackage{wrapfig}
\usepackage{relsize}
\usepackage{tabularx}
\usepackage{array}
\usepackage[accsupp]{axessibility}  %
\newsavebox\CBox

\usepackage{xcolor}
\usepackage{overpic}
\usepackage{colortbl}

\usepackage{graphicx}

\usepackage{caption}

\graphicspath{{images/}}
\newcommand{\tabref}[1]{Tab.~\ref{#1}}

\newcommand{\equref}[1]{Eq.~\ref{#1}}
\renewcommand{\figref}[1]{Fig.~\ref{#1}}
\renewcommand{\secref}[1]{Sec.~\ref{#1}}
\usepackage{amsmath}
\usepackage{dsfont}
\usepackage{graphicx} %
\usepackage{caption}  %
\usepackage{capt-of}  %
\usepackage{enumitem}
\usepackage{booktabs}
\usepackage{graphicx} %
\usepackage{caption} %
\usepackage{pifont} %
\usepackage{threeparttable}

\title{SpatialSpeak: QA-Native Reconstruction with Local and Global Context for Spatial Chain-of-Thought Reasoning}

\author{%
Yang Cao$^{1}$ \quad
Jiaxin Zhang$^{3}$ \quad
Dave Zhenyu Chen$^{2}$ \quad
Yingji Zhong$^{1}$ \\[0.25em]
\bfseries Ruiyuan Gao$^{2}$ \quad
Lanqing Hong$^{2}$ \quad
Dan Xu$^{1}$\thanks{Corresponding author}
  \vspace{0.1cm}
  \\
  $^{1}$Hong Kong University of Science and Technology \\
  $^{2}$Huawei Noah’s Ark Lab \quad
  $^{3}$Harbin Institute of Technology
}

\hypersetup{
  hidelinks,
  pdftitle={SpatialSpeak: QA-Native Reconstruction with Local and Global Context for Spatial Chain-of-Thought Reasoning},
  pdfauthor={Yang Cao, Jiaxin Zhang, Dave Zhenyu Chen, Yingji Zhong, Ruiyuan Gao, Lanqing Hong, Dan Xu}
}
\begin{document}

\maketitle
\fancyhead{}
\lhead{Preprint}
\renewcommand{\headrulewidth}{0pt}

\begin{abstract}

Vision-language models (VLMs) can benefit from geometric priors for multi-view spatial reasoning, yet answer-only training does not directly supervise the intermediate geometric estimates and their use in deriving quantitative spatial answers. We hypothesize that spatial chain-of-thought (CoT) supervision becomes more effective when the VLM first jointly learns complementary local geometry and global scene context through multi-view reconstruction. We introduce \textbf{SpatialSpeak}, a two-stage framework that connects QA-native reconstruction pretraining with spatial CoT learning. In Stage~I, QA-Native Reconstruction Pretraining (QA-RP) combines marked-point 3D queries for fine-grained local geometry with object-center queries for global scene context across views. Both tasks are formulated as text-based question answering, allowing geometric estimation and subsequent reasoning to share the same autoregressive output interface. In Stage~II, spatial CoT with Visual Compensation (CoT-VC) trains the model to express question-relevant geometric estimates and use them to derive answers, with reliability assessment and visual compensation supporting answer refinement when needed. On ReVSI, QA-RP increases the gain from CoT-VC from 2.6 to 6.9 points, and ablations show that both local and global reconstruction supervision are beneficial. SpatialSpeak achieves state-of-the-art results on ReVSI, VSI-Bench, and SPAR-Bench, with a ReVSI score of 62.8 that exceeds the strongest compared baseline by 8.7 points. Project page: https://yangcaoai.github.io/SpatialSpeak/.

\end{abstract}

\begin{figure}[t!]
     \centering
    \begin{overpic}[width=1\columnwidth]
    {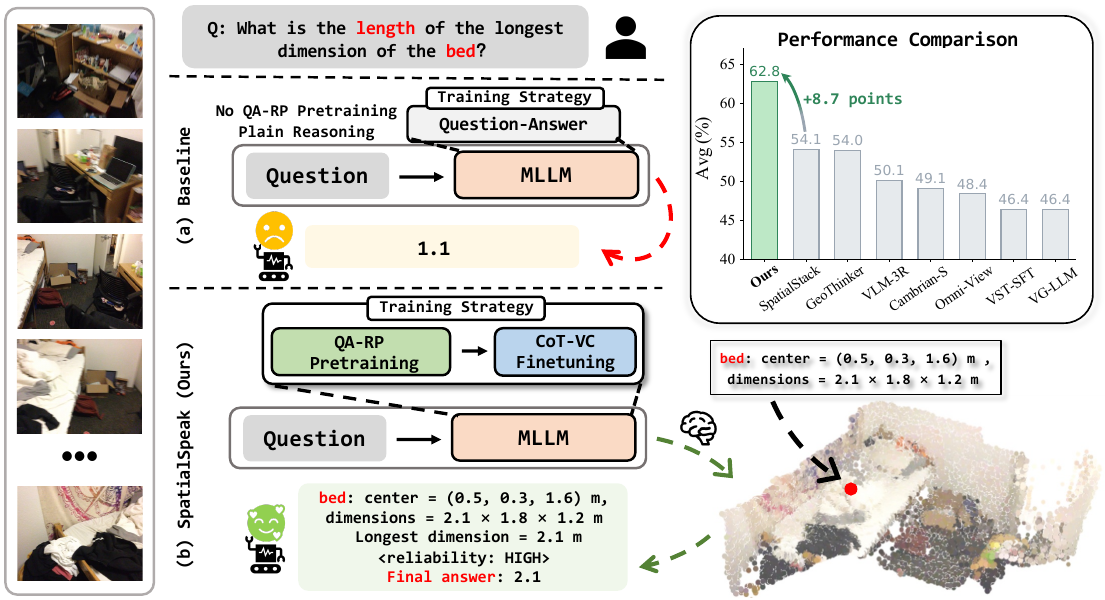}
    \end{overpic}
    \vspace{-0.6cm}    
    \caption{
Panel (a) shows our baseline, which incorporates geometric priors through feature fusion and is trained on spatial-reasoning QA without QA-RP or CoT-VC. Panel (b) shows \textbf{SpatialSpeak}, which trains the VLM to conduct multi-view 3D reconstruction through QA-Native Reconstruction Pretraining (QA-RP) with complementary local geometry and global scene context, then to involve the learned geometry in explicit spatial reasoning through spatial Chain-of-Thought with Visual Compensation~(CoT-VC), achieving leading results on ReVSI~\citep{zhang2026revsi}.
The top-right chart compares average ReVSI scores with SpatialStack-4B~\citep{zhang2026spatialstack}, GeoThinker-8B~\citep{li2026thinking}, VLM-3R-7B~\citep{fan2025vlm}, Cambrian-S-7B~\citep{yang2025cambrians}, Omni-View-7B~\citep{hu2025omniview}, VST-7B-SFT~\citep{vst}, and VG-LLM-8B~\citep{zhenglearning}. SpatialSpeak-4B achieves 62.8, outperforming SpatialStack-4B (54.1) by 8.7 points.
}
\vspace{-0.4cm}
\label{fig:teaser}
\end{figure}

\section{Introduction}

Multi-view vision-language models (VLMs)~\citep{wu2025spatialmllm,tong2024cambrian,huang2025,li2026thinking}  are increasingly equipped with 3D geometric priors~\citep{wang2025continuous,wang2025vggt,wang2025flashvggt} for spatial question answering. 
Existing approaches commonly inject features from pretrained reconstruction models into VLMs~\citep{fan2025vlm,zhenglearning,zhang2026spatialstack}, a paradigm adopted by our baseline~(\figref{fig:teaser}a). Reconstruction-~\citep{hu2025g} and generation-based~\citep{hu2025omniview} objectives further improve scene representations. These advances strengthen spatial understanding~\citep{yang2025thinking,cao2023coda,cao2026vggtdet,fan2024largespatialmodelendtoend,yan2025pgocc}, but geometric learning alone does not directly teach a VLM how to express and use question-relevant geometric estimates within explicit reasoning traces.

Answer-only supervision leaves this reasoning process implicit: it provides no direct supervision for the intermediate geometric estimates and their use to derive spatial answers. This limitation is particularly important for quantitative spatial questions, for which reliable answers depend on both accurate geometric estimation and appropriate geometric reasoning. We therefore introduce spatial chain-of-thought~(CoT) supervision, which explicitly specifies question-relevant geometric estimates and the steps used to derive the final answer. We further hypothesize that such supervision is more effective when the VLM first jointly learns complementary local geometry and global scene context through multi-view reconstruction. Local reconstruction grounds image points in 3D, whereas global reconstruction captures the spatial arrangement of object instances across views.

Based on this insight, we introduce \textbf{SpatialSpeak}, a two-stage learning framework that combines QA-Native Reconstruction Pretraining~(QA-RP) with spatial Chain-of-Thought with Visual Compensation~(CoT-VC) supervision (Fig.~\ref{fig:teaser}). In Stage I, QA-RP jointly supervises local point reconstruction and global object-center reconstruction. Local queries ask the VLM to predict the 3D location of a marked image point from multi-view inputs, providing fine-grained geometric supervision. Global queries ask it to enumerate visible object instances and predict their semantic categories and 3D centers in a shared coordinate system, providing scene-wide object-layout supervision. Crucially, both tasks are formulated as text-based question answering and trained with the standard next-token prediction objective. This QA-native design aligns geometric prediction with the autoregressive output interface later used for spatial reasoning.

In Stage II, CoT-VC trains the pretrained VLM to identify question-relevant entities, express their geometric estimates, and derive answers through explicit task-specific computations. Because geometric derivations can be approximate, CoT-VC also supervises reliability assessment and answer refinement using direct visual evidence when needed. On ReVSI, QA-RP increases the gain from CoT-VC from 2.6 to 6.9 points, and ablating either local or global supervision degrades performance. SpatialSpeak achieves state-of-the-art results on ReVSI~\citep{zhang2026revsi}, VSI-Bench~\citep{yang2025thinking} and SPAR-Bench~\citep{zhang2025from}.

Our contributions are as follows: 

\begin{itemize}[leftmargin=*]

\item We establish that multi-view reconstruction pretraining substantially increases the benefit of explicit spatial reasoning supervision: the gain from CoT-VC rises from 2.6 points without QA-RP to 6.9 points with QA-RP on ReVSI. SpatialSpeak achieves state-of-the-art results on ReVSI, VSI-Bench, and SPAR-Bench.

\item We propose QA-Native Reconstruction Pretraining (QA-RP), which learns complementary local point geometry and global object-layout context through text-based QA targets in a shared 3D coordinate system.

 \item We introduce spatial Chain-of-Thought with Visual Compensation (CoT-VC), which explicitly supervises question-relevant geometric estimates, task-specific answer derivations, reliability assessment, and visually grounded answer refinement. 

\end{itemize}

\vspace{-0.2cm}

\section{Related Work}
\paragraph{3D Reconstruction.}
Given multi-view images without known poses, 3D reconstruction~\citep{hartley2003multiple,leroy2024grounding,zhong2025empowering,mione4d2025} seeks to recover both the scene’s geometry and the associated camera trajectories. Traditional pipelines split this objective into modules such as multi-view stereo~\citep{wang2021patchmatchnet,zhang2020visibility}, feature matching~\citep{lindenberger2023lightglue,sun2021loftr}, and keypoint detection~\citep{lowe2004distinctive,detone2018superpoint}, among others. More recently, DUSt3R~\citep{wang2024dust3r} proposed a unified network that predicts scene structure directly, shifting the conventional paradigm. Then
MASt3R~\citep{leroy2024grounding} adds an auxiliary head for correspondence estimation. However, both DUSt3R and MASt3R process only image pairs, which restricts global context, requires multiple forward passes, and entails an expensive global alignment stage. Fast3R~\citep{yang2025fast3r} mitigates these issues by consuming long sequences in a single pass and eliminating coordinate alignment. 
InstantSplat~\citep{fan2024instantsplat} is a fast, self-supervised method that uses Gaussian Bundle Adjustment  and co-visibility to jointly recover geometry from  2-3 unposed views.
CUT3R~\citep{wang2025continuous} introduces a recurrent transformer that incrementally produces a unified, metric-scale reconstruction from image streams. 
Beyond point maps, VGGT~\citep{wang2025vggt} further estimates camera poses and other 3D  attributes. DepthLM~\citep{cai2025depthlm} demonstrates expert-level single-image metric depth estimation with text-based SFT. It also extends to two-point distance estimation and camera displacement estimation from image pairs. Our focus is on learning multi-view geometry with local and global context to support spatial CoT.

\paragraph{Spatial MLLMs.} Endowing multimodal large language models (MLLMs), also called vision-language models (VLMs)~\citep{singh2025openai,team2023gemini,Qwen3-VL,zheng2024video3dllm,xu2025uniugg,li2025does,wang2025n3d,zhu2026partllm}, with fine-grained spatial understanding has attracted growing attention. A dominant paradigm relies on pretrained feed-forward 3D reconstruction models~\citep{wang2025vggt,wang2025continuous,wang2024dust3r} as external geometry providers and injects their reconstruction prior into the language model's representation space. Along this line, \emph{token-level feature fusion} is a common mechanism. VG-LLM~\citep{zhenglearning} directly adds geometric tokens to visual tokens. Spatial-MLLM~\citep{wu2025spatialmllm} pairs a 2D semantic visual encoder with a geometry-prior spatial encoder and space-aware frame sampling to improve spatial reasoning. VLM-3R~\citep{fan2025vlm} applies cross-attention layers that allow visual representations to query geometric tokens. GeoThinker~\citep{li2026thinking} further shifts from passive fusion to selective integration of geometric evidence across multiple levels. Beyond feature integration, other approaches incorporate geometric training objectives. GAP-MLLM~\citep{zhang2026gapmllm} combines semantic labeling and sparse 3D point prediction with multi-level gated fusion to improve downstream 3D perception. G$^2$VLM~\citep{hu2025g} integrates geometric and semantic experts through shared self-attention and learns scene geometry with dedicated prediction heads. In its released question-answering implementation, the semantic branch conditions answer generation on the geometric expert's hidden representations through attention. Omni-View~\citep{hu2025omniview} jointly trains 3D scene understanding, novel-view synthesis, and depth and camera-pose estimation, using dedicated texture and geometry modules to strengthen scene understanding.
Another strategy is \emph{feature distillation} or alignment. 3DRS~\citep{huang2025} transfers knowledge from a frozen reconstruction model into the visual encoder, while Spatial Forcing~\citep{li2025spatial} enforces direct embedding alignment between visual and geometric streams during training. Beyond token fusion and distillation, SpatialStack~\citep{zhang2026spatialstack} aligns and stacks multi-scale geometric features with the language backbone, and Map2Thought~\citep{gao2026map2thought} leverages external metric-scale scene graphs to facilitate spatial reasoning.
Our work investigates how jointly learning local point geometry and global scene context through QA-based reconstruction supervision supports subsequent spatial CoT learning. We explicitly supervise how question-relevant geometric estimates are expressed and used to derive answers, with visual compensation supporting refinement when needed.

\section{Method} \label{sec:method}
\begin{figure}[t!]
     \centering
    \begin{overpic}[width=1\columnwidth]
    {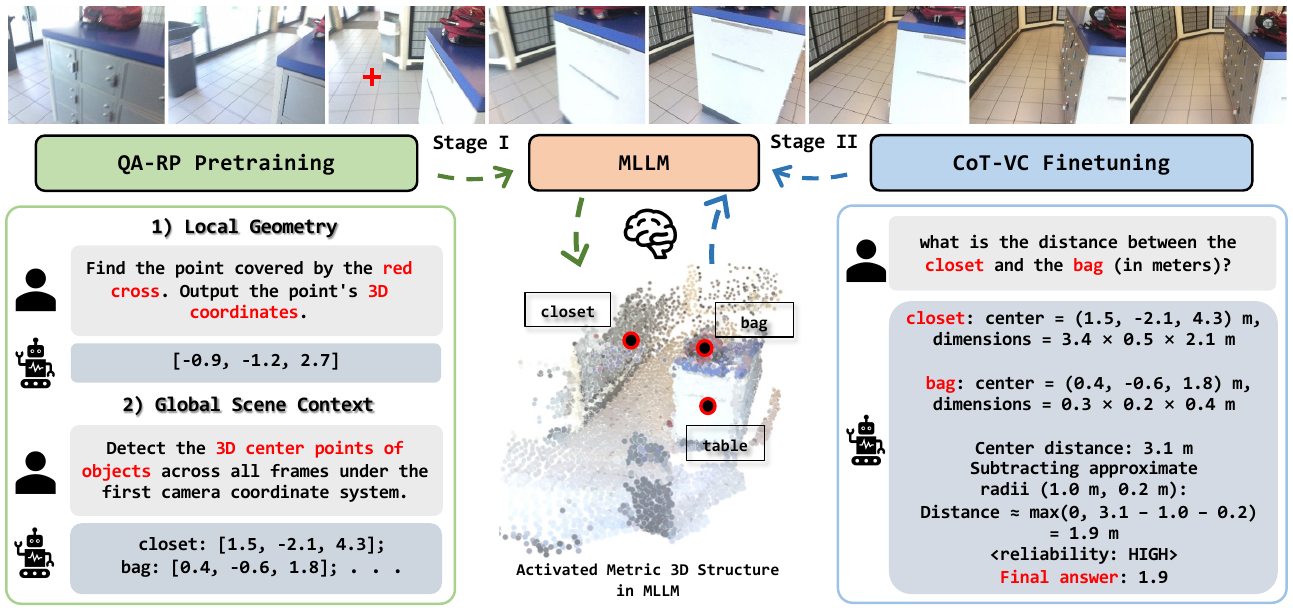}
    \end{overpic}
    \vspace{-0.6cm}  
    \caption{
\textbf{Overview of the SpatialSpeak framework}. In Training Stage~I, QA-Native Reconstruction Pretraining (QA-RP) trains the VLM to conduct metric-scale multi-view reconstruction within the standard QA interface. Local point queries and global object-center queries jointly supervise fine-grained local geometry and global scene context, with both targets expressed in the first-frame camera coordinate system. In Training Stage~II, spatial Chain-of-Thought with Visual Compensation (CoT-VC) trains the model to reason explicitly with the learned geometry, assess the reliability of geometry-derived answers, and refine answers using direct visual evidence when needed.}
\vspace{-0.5cm}
\label{fig:method}
\end{figure}

In this section, we present our approach in detail. The examples in \figref{fig:teaser} and \figref{fig:method} are constructed to show the workflow.
We begin with \textit{Preliminaries} (\secref{sec:prelim}),
introducing our baseline and the notation for its inputs and outputs.
We then describe our two-stage reconstruction-to-reasoning learning framework:
\textit{Stage~I: QA-Native Reconstruction Pretraining with Local and Global Context} (\secref{sec:stage1}),
which learns complementary local geometry and global scene context through QA-native reconstruction supervision,
followed by \textit{Stage~II: Spatial Chain-of-Thought with Visual Compensation} (\secref{sec:stage2}),
which trains the VLM to express geometric estimates and use them to derive spatial answers,
with visual compensation supporting answer refinement when needed.  

\subsection{Preliminaries}
\label{sec:prelim}
Our baseline~(`MLLM' in ~\figref{fig:method}) is built on the popular VG-LLM~\citep{zhenglearning},
which takes as input a sequence of $N$ RGB frames
$\mathcal{I} = \{I_1, \ldots, I_N\}$ together with a natural-language query $q$
and produces a text response $y$ via autoregressive decoding.

\textbf{Feature Encoding.}
Each frame $I_i$ is first processed by a vision encoder $\phi_\text{vis}$
to obtain a sequence of patch tokens:
\begin{equation}
  \mathbf{v}_i = \phi_\text{vis}(I_i) \in \mathbb{R}^{L \times d_v},
  \quad i = 1,\ldots,N,
\end{equation}
where $L$ is the number of patch tokens per frame and $d_v$ is the visual feature dimension.
In parallel, a frozen multi-view geometry encoder $\phi_\text{geo}$ of VGGT~\citep{wang2025vggt}
processes all $N$ frames jointly and extracts 3D-aware features, which are then fused with visual features,
combining geometry priors for the reasoning.
Let $\tilde{\mathbf{v}}_i$ denote the geometry-fused visual tokens.

\textbf{Language Modeling.}
The text query is tokenized into $\mathbf{q} \in \mathbb{Z}^M$.
The LLM $f_\theta$ receives the concatenation of all projected visual tokens
and query tokens and produces the response via:
\begin{equation}
  y = f_\theta\!\left(\bigl[\phi_\text{proj}(\tilde{\mathbf{v}}_1),\, \ldots,\, \phi_\text{proj}(\tilde{\mathbf{v}}_N),\, \mathbf{q}\bigr]\right),
\end{equation}
where $\phi_\text{proj}$ is a learned MLP projector aligning visual tokens to the LLM's
hidden dimension.
We optimize only the LLM parameters with a next-token prediction loss
over the target response tokens, keeping the vision encoder, geometry encoder,
and projector frozen.

\subsection{Stage~I: QA-Native Reconstruction Pretraining with Local and Global Context}
\label{sec:stage1}

To prepare the VLM for spatial CoT reasoning, Stage I jointly learns complementary local geometry and global scene context through multi-view reconstruction. Local point queries ground marked image points in 3D, while global object-center queries ask the model to enumerate object instances and predict their semantic categories and 3D centers across views. All predicted 3D coordinates are expressed in the first-frame camera coordinate system, giving points and objects a shared spatial reference. Both tasks are formulated as text-based QA, allowing geometric prediction and subsequent spatial reasoning to share the same autoregressive output interface.

\paragraph{Local Reconstruction Task.}
As shown in `QA-RP Pretraining' of~\figref{fig:method}, to ground fine-grained local geometry in explicit 3D coordinates, we supervise
local point-wise 3D reconstruction inspired by prior work~\citep{cai2025depthlm,zhang2026gapmllm}.
For each training sample we select a scene with $N$ video frames
$\{I_1,\ldots,I_N\}$ and mark a 2D point $(u, v)$ in frame $I_i$ with a red cross.
The local query $q_\text{local}$ asks for the point's 3D coordinates
in the camera coordinate system of the first frame:
\begin{quote}
``Find the point covered by the red cross. Output the point's 3D coordinates.''
\end{quote}
The ground-truth response is the lifted 3D point
$\mathbf{p}^{(1)} = [x,\, y,\, z]^\top$ expressed in the first-frame camera coordinate system following VGGT~\citep{wang2025vggt},
computed from the ScanNet depth maps and known camera extrinsics:
\begin{equation}
  \mathbf{p}^{(1)} = \mathbf{R}_{i\to 1}\,\mathbf{p}^{(i)} + \mathbf{t}_{i\to 1},
\end{equation}
where $\mathbf{R}_{i\to 1},\,\mathbf{t}_{i\to 1}$ are the relative rotation and translation from frame $i$ to frame $1$.
The model outputs the prediction:
$\hat{y}_\text{local} = $ \texttt{\{[$\hat{x}$, $\hat{y}$, $\hat{z}$]\}}.

\begin{figure}[t!]
     \centering
    \begin{overpic}[width=1\columnwidth]
    {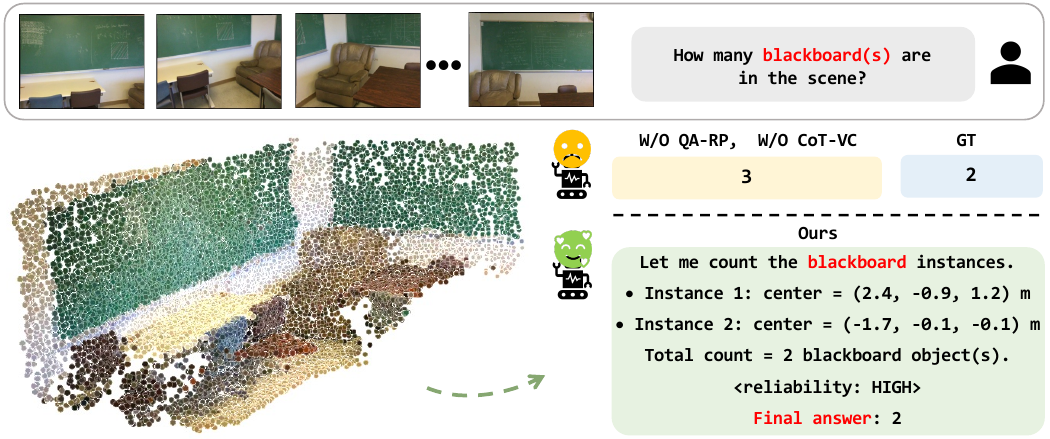}
    \end{overpic}
    \vspace{-0.6cm}  
    \caption{
\textbf{Qualitative example}. The bottom-left panel shows the scene reconstruction predicted after Stage~I (QA-RP), and the bottom-right panel shows the spatial CoT output after Stage~II (CoT-VC). SpatialSpeak lists two blackboard instances with their estimated 3D centers and predicts a count of 2, matching the ground truth. The variant with neither QA-RP nor CoT-VC predicts 3.}
\vspace{-0.5cm}
\label{fig:demov2}
\end{figure}

\paragraph{Global Reconstruction Context.}
While the local task supervises a single 3D point,
it does not explicitly cover the distribution of objects across the scene.
We therefore introduce a complementary \emph{global} task that requires enumerating
all visible object instances across all $N$ frames.
Given the multi-frame input, the global query $q_\text{global}$ is:
\begin{quote}
``Detect the 3D center points of objects across all frames
under the first frame coordinate system.''
\end{quote}
The ground-truth response is a list of object detections,
each consisting of a semantic label $c_k$ and a 3D center
$\mathbf{c}_k^{(1)} \in \mathbb{R}^3$ transformed to the first-frame coordinate system.
Note that objects are ordered by their first appearance (earlier frames first) and,
within the same frame, by spatial location (left-to-right, top-to-bottom),
to produce a deterministic, perceptually natural output sequence.
This global task complements local point supervision with the semantic categories and spatial arrangement of scene objects in the same reference coordinate system.

\paragraph{Joint Supervision.}
We train the model on a mixture of local and global reconstruction samples
using the standard next-token prediction objective:
\begin{equation}
  \mathcal{L}_\text{Stage\,I}
  = \mathbb{E}_{\mathcal{D}_\text{local}}\bigl[-\log p_\theta(y_\text{local}\mid \mathcal{I},q_\text{local})\bigr]
  + \mathbb{E}_{\mathcal{D}_\text{global}}\bigl[-\log p_\theta(y_\text{global}\mid \mathcal{I},q_\text{global})\bigr].
\end{equation}
Both geometric targets are learned through text responses, without auxiliary 3D regression heads or additional geometric losses.

\begin{table}[!t]
\centering
\captionsetup{font=small,skip=5pt,justification=justified,singlelinecheck=false}
\begin{minipage}[t]{0.32\linewidth}
\vspace{0pt}
\centering
\small
\setlength{\tabcolsep}{4pt}
\caption{\textbf{Training components}. QA-RP and CoT-VC each improve spatial reasoning, with larger gains when both designs are used.}
\vspace{-0.1cm}
\renewcommand{\arraystretch}{1.1}
\begin{tabularx}{\linewidth}{>{\raggedright\arraybackslash}Xc}
\specialrule{\lightrulewidth}{\abovetopsep}{0pt}
\noalign{\begingroup\color{orange!10}\hrule height \belowrulesep\endgroup}
\rowcolor{orange!10}
Method   & Avg.   \\
\noalign{\begingroup\color{orange!10}\hrule height \aboverulesep\endgroup}
\specialrule{\lightrulewidth}{0pt}{\belowrulesep}

\model~(full) & \textbf{62.8} \\ %
w/o CoT-VC  & 55.9  \\ %
w/o QA-RP  & 55.0  \\ %
w/o both  & 52.4 \\ %
\bottomrule[\lightrulewidth]
\end{tabularx}
\label{tab:ablation1}
\end{minipage}
\hfill
\begin{minipage}[t]{0.32\linewidth}
\vspace{0pt}
\centering
\small
\setlength{\tabcolsep}{4pt}
\caption{\textbf{QA-RP context}. Ablating global object-center queries, local point queries, or both from multi-view reconstruction pretraining.}
\vspace{-0.1cm}
\renewcommand{\arraystretch}{1.1}
\begin{tabularx}{\linewidth}{>{\raggedright\arraybackslash}Xc}
\specialrule{\lightrulewidth}{\abovetopsep}{0pt}
\noalign{\begingroup\color{orange!10}\hrule height \belowrulesep\endgroup}
\rowcolor{orange!10}
Method   & Avg.   \\
\noalign{\begingroup\color{orange!10}\hrule height \aboverulesep\endgroup}
\specialrule{\lightrulewidth}{0pt}{\belowrulesep}

\model~(full)  & \textbf{62.8} \\ %
w/o global queries     & 59.9  \\ %

w/o local queries      & 59.3 \\ %
w/o both   & 55.0  \\ %

\bottomrule[\lightrulewidth]
\end{tabularx}
\label{tab:ablation2}
\end{minipage}
\hfill
\begin{minipage}[t]{0.32\linewidth}
\vspace{0pt}
\centering
\small
\setlength{\tabcolsep}{4pt}
\renewcommand{\arraystretch}{1.1}
\caption{\textbf{Reliability threshold}. Varying $\tau$ for reliability labels in CoT-VC (\secref{sec:stage2}). All three outperform the variant w/o CoT-VC.}
\vspace{-0.1cm}
\begin{tabularx}{\linewidth}{>{\raggedright\arraybackslash}Xc}
\specialrule{\lightrulewidth}{\abovetopsep}{0pt}
\noalign{\begingroup\color{orange!10}\hrule height \belowrulesep\endgroup}
\rowcolor{orange!10}
Method    & Avg.   \\
\noalign{\begingroup\color{orange!10}\hrule height \aboverulesep\endgroup}
\specialrule{\lightrulewidth}{0pt}{\belowrulesep}

\model~($\tau=0.1$)  & 59.6  \\ %
 \model~($\tau=0.3$)  & \textbf{62.8} \\ %
\model~($\tau=0.5$) &  59.5  \\ %
w/o CoT-VC  & 55.9  \\ %
\bottomrule[\lightrulewidth]
\end{tabularx}
\label{tab:reliability_threshold}
\end{minipage}

\vspace{-0.5cm}
\end{table}

\subsection{Stage~II: Spatial Chain-of-Thought with Visual Compensation}
\label{sec:stage2}

Stage~I supervises scene geometry, but its reconstruction targets do not specify how to derive answers to spatial questions. Stage~II therefore complements reconstruction pretraining with spatial CoT supervision that explicitly connects question-relevant geometric estimates to answer derivation. As shown in `CoT-VC Finetuning' of~\figref{fig:method}, we construct structured responses that identify relevant objects, state their geometric estimates, and derive answers through explicit computations. We further include Visual Compensation (VC) to supervise reliability assessment and answer refinement using visual evidence when needed, yielding CoT-VC.

\paragraph{Spatial CoT Data Construction.}
Each CoT training sample in Stage~II is a spatial QA pair
$(q, a^\star)$ drawn from ScanNet~\citep{dai2017scannet} following VLM-3R~\citep{fan2025vlm} enriched with 3D object annotations
(semantic labels, 3D bounding boxes).
We implement deterministic CoT generators for four question types:
\emph{distance}, \emph{size}, \emph{count}, and \emph{closest object}.
For each type we derive a 3D estimate $\hat{a}$ from the annotations
and compare it with the ground-truth answer $a^\star$ to assign a reliability label.

\begin{itemize}[leftmargin=1.5em]
\item \textbf{Distance.}
  Given two object categories $(c_1, c_2)$,
  we select the most-visible instance of each
  (the one appearing in the most frames)
  and compute an approximate closest-point distance as
  \begin{equation}
    d_\text{cp} = \max\!\bigl(0,\; \|\mathbf{c}_1^{(1)} - \mathbf{c}_2^{(1)}\|_2 - r_1 - r_2\bigr),
    \label{eq:dist}
  \end{equation}
  where $r_k = \tfrac{1}{6}(s_k^x + s_k^y + s_k^z)$ is
  the mean half-side of object $k$'s axis-aligned bounding box, used as an approximate radius.
  The estimate is labeled \textsc{High} if
  $|d_\text{cp} - a^\star| / a^\star \le \tau$; otherwise \textsc{Low}.

\item \textbf{Size.}
  We obtain the longest side length of the most-visible instance's bounding box,
  $\ell = \max(s^x, s^y, s^z)$, converted to centimeters.
  Reliability threshold: relative error $\le \tau$.

\item \textbf{Count.}
  We count the number of detected instances of category $c$.
  The estimate is \textsc{High} if and only if it equals the integer ground truth.

\item \textbf{Closest Object.}
  For a multiple-choice question, we compute the approximate distance $d_\text{cp}$ from~\equref{eq:dist} for each option
  and predict the letter corresponding to the closest option.
  Reliability is \textsc{High} if the predicted letter matches $a^\star$.
\end{itemize}

\paragraph{Spatial CoT Template.}
Because geometry-based computations can involve approximations, the derived answer need not always agree with the QA ground truth. We therefore include reliability-conditioned refinement in the CoT target, retaining the geometric derivation while allowing the final answer to be revised using visual evidence. Specifically, each response contains the 3D reasoning chain followed by a reliability token:
\begin{equation}
y_\text{CoT} =
\underbrace{\mathcal{C}_\text{3D}}_{\text{3D reasoning chain}}
\oplus
\begin{cases}
[\textsc{High}] \oplus \hat{a}
  & \text{if } \hat{a} \text{ is reliable,}\\[6pt]
[\textsc{Low}]  \oplus \mathcal{V} \oplus a^\star
  & \text{otherwise,}
\end{cases}
\label{eq:cot}
\end{equation}
where $\mathcal{C}_\text{3D}$ is the 3D reasoning chain always present in the response,
$[\textsc{High}]$ / $[\textsc{Low}]$ are the reliability tokens,
$\hat{a}$ is the 3D-derived answer, $\mathcal{V}$ is the visual refinement note,
and $a^\star$ is the ground-truth answer used as supervision.
When the 3D estimate is reliable (\textsc{High}), the model follows the geometric chain and outputs the 3D-derived answer. When it is not (\textsc{Low}), the model is trained to acknowledge the limitation, invoke a visual-inspection fallback (``\emph{refining based on visual observation of the scene}''), and output the ground-truth answer. This supervision trains the model to assess its geometric estimates and to refine the final answer with visual evidence when needed.

\paragraph{Stage~II Training.}
The model is initialized from the Stage~I checkpoint and fine-tuned on the
spatial CoT dataset $\mathcal{D}_\text{CoT}$
using the same next-token prediction loss:
\begin{equation}
  \mathcal{L}_\text{Stage\,II}
  = \mathbb{E}_{(q, y_\text{CoT})\sim \mathcal{D}_\text{CoT}}
  \bigl[-\log p_\theta(y_\text{CoT} \mid \mathcal{I}, q)\bigr].
\end{equation}
For question types not covered by these geometric CoT templates (\emph{e.g.}, relative direction),
we retain direct answer supervision to preserve coverage of the full VSI-Bench task distribution.

\begin{table}[!t]
\centering
\captionsetup{font=small,skip=5pt,justification=justified,singlelinecheck=false}
\begin{minipage}[t]{0.32\linewidth}
\vspace{0pt}
\centering
\small
\setlength{\tabcolsep}{4pt}
\renewcommand{\arraystretch}{1.1}
\caption{\textbf{Spatial CoT}. All model variants retain QA-RP. Removing VC preserves spatial CoT. Removing CoT-VC retains only direct answer supervision in Stage~II.}
\vspace{-0.1cm}
\begin{tabularx}{\linewidth}{>{\raggedright\arraybackslash}Xc}
\specialrule{\lightrulewidth}{\abovetopsep}{0pt}
\noalign{\begingroup\color{orange!10}\hrule height \belowrulesep\endgroup}
\rowcolor{orange!10}
Method  &  Avg.   \\
\noalign{\begingroup\color{orange!10}\hrule height \aboverulesep\endgroup}
\specialrule{\lightrulewidth}{0pt}{\belowrulesep}

\model~(full)  & \textbf{62.8} \\
w/o VC    & 58.5   \\ %
w/o CoT-VC   & 55.9    \\ %

\bottomrule[\lightrulewidth]
\end{tabularx}
\label{tab:ablation3}
\end{minipage}
\hfill
\begin{minipage}[t]{0.66\linewidth}
\vspace{0pt}
\centering
\small
\setlength{\tabcolsep}{4pt}
\renewcommand{\arraystretch}{1.1}
\caption{\textbf{Pointmap reconstruction on ScanNet}. Acc. and Comp. are evaluated without alignment. Acc.$^{*}$ and Comp.$^{*}$ are evaluated after GT Sim(3) alignment. All errors are reported in cm. Lower values are better. Baselines are MapAnything~\citep{keetha2026mapanything} and CUT3R~\citep{wang2025continuous}. Bold type marks the lowest error in each column.}
\vspace{-0.1cm}
\begin{tabularx}{\linewidth}{l*{4}{>{\centering\arraybackslash}X}}
\specialrule{\lightrulewidth}{\abovetopsep}{0pt}
\noalign{\begingroup\color{orange!10}\hrule height \belowrulesep\endgroup}
\rowcolor{orange!10}
Method & Acc. $\downarrow$ & Comp. $\downarrow$ & Acc.$^{*}$ $\downarrow$ & Comp.$^{*}$ $\downarrow$ \\
\noalign{\begingroup\color{orange!10}\hrule height \aboverulesep\endgroup}
\specialrule{\lightrulewidth}{0pt}{\belowrulesep}
MapAnything & 36.3 & 28.4 & 5.7 & 6.2 \\
CUT3R & 13.7 & 12.7 & \textbf{4.7} & \textbf{4.6} \\

\model~(Ours) & \textbf{8.9} & \textbf{9.3} & 5.0 & 5.0 \\
\bottomrule[\lightrulewidth]
\end{tabularx}
\label{tab:ablation5}
\end{minipage}

\vspace{-0.15cm}
\end{table}

\begin{figure}[t!]
     \centering
    \begin{overpic}[width=1\columnwidth]
    {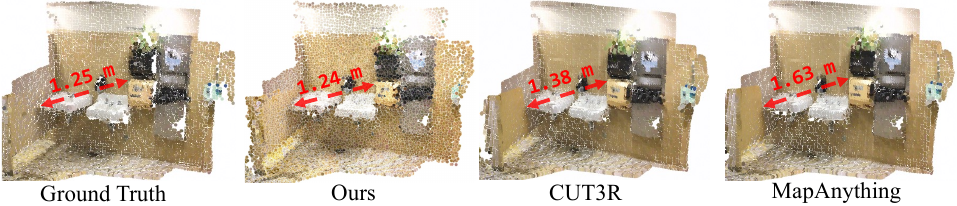}
    \end{overpic}
    \vspace{-0.6cm}
    \caption{
\textbf{Qualitative reconstruction comparison}. Red dashed arrows mark corresponding distances in meters. SpatialSpeak estimates the distance as 1.24~m, close to the ground truth of 1.25~m.}
\vspace{-0.5cm}

\label{fig:recons_demo1}
\end{figure}

\section{Experiments} \label{sec:experiments}

\paragraph{Datasets and Benchmarks.}
Training Stage~I uses the reconstruction pretraining data described in~\secref{sec:stage1}.
For Training Stage~II, following VG-LLM~\citep{zhenglearning}, we use the same training subsets from the LLaVA-Hound split of LLaVA-Video-178K~\citep{zhang2024llava} and SPAR-7M~\citep{zhang2025from}, augmented with our spatial
CoT data~(\secref{sec:stage2}).
We evaluate spatial reasoning on ReVSI~\citep{zhang2026revsi}, VSI-Bench~\citep{yang2025thinking}, and SPAR-Bench~\citep{zhang2025from}.
ReVSI corrects annotation errors and reduces answer-distribution bias in VSI-Bench, and is used for all spatial-reasoning ablations.
Metric-scale reconstruction is evaluated on ScanNet.

\paragraph{Implementation Details.}
We initialize from the popular Qwen3-VL-4B~\citep{Qwen3-VL} and fine-tune only the LLM parameters
while keeping both the vision encoder and the MLP projector frozen.
All stages share the same base configuration: a cosine learning rate schedule
with a warm-up ratio of 0.03, weight decay of $0.01$, and 32 frames sampled per video clip.
Stage~I trains for 1 epoch with a global batch size of 32
and a peak learning rate of $5\times10^{-6}$.
Stage~II trains for 1 epoch with a global batch size of 64
and a peak learning rate of $1\times10^{-5}$. More details are
given in Appendix~\secref{sec:app_imple}.

\subsection{Ablation Study}
We conduct spatial-reasoning ablations on ReVSI and report the average score across its seven task categories.
We also evaluate Stage~I reconstruction on ScanNet.
Throughout the ablations, `w/o CoT-VC' retains Stage~II training with answer-only supervision.

\paragraph{Effectiveness of Key Components.}
\tabref{tab:ablation1} reports the contribution of each component on ReVSI. The baseline without either component scores 52.4. QA-RP alone raises the score to 55.9, while CoT-VC alone raises it to 55.0. Combining both achieves 62.8, an improvement of 10.4 points over the baseline. Removing QA-RP or CoT-VC from the full method reduces the score by 7.8 or 6.9 points, respectively. The gain from CoT-VC increases from 2.6 without QA-RP to 6.9 points with QA-RP. The larger gain from CoT-VC after QA-RP supports our hypothesis that jointly learning local geometry and global scene context provides a stronger foundation for spatial CoT learning.

\paragraph{Analysis of Reconstruction Pretraining.}
\tabref{tab:ablation2} examines the local and global supervision in QA-RP. The full method scores 62.8, compared with 59.9 when global object-center queries are removed and 59.3 when local point queries are removed. Removing both reduces the score to 55.0. 
These results support jointly learning local geometry and global scene context during reconstruction pretraining to better prepare the model for subsequent spatial reasoning.

\paragraph{Analysis of the Reliability Threshold.}
\tabref{tab:reliability_threshold} evaluates the relative-error threshold $\tau$ used to construct reliability labels. Among the tested values, $\tau=0.3$ achieves the highest score of 62.8, compared with 59.6 for $\tau=0.1$ and 59.5 for $\tau=0.5$. All three settings outperform the variant without CoT-VC (55.9).

\paragraph{Analysis of Spatial CoT with Visual Compensation.}
\tabref{tab:ablation3} examines the roles of spatial CoT and VC in Stage~II. Starting from the model without CoT-VC, which scores 55.9, adding spatial CoT without VC improves the score to 58.5. Incorporating VC further raises it to 62.8, an additional gain of 4.3 points. 
These results support explicitly training the model both to reason with geometric estimates and to refine its answers when needed.

\paragraph{3D Pointmap Reconstruction Accuracy.}
In~\tabref{tab:ablation5}, we compare our Stage~I model with CUT3R and MapAnything on the same 200 randomly sampled ScanNet validation scenes held out from training, under two protocols: no alignment and GT Sim(3) alignment. Without alignment, SpatialSpeak achieves the lowest Acc. and Comp. errors (8.9 and 9.3~cm), compared with CUT3R (13.7 and 12.7~cm) and MapAnything (36.3 and 28.4~cm). After GT Sim(3) alignment, SpatialSpeak obtains 5.0~cm on both metrics,  lower than MapAnything (5.7 and 6.2~cm) but slightly higher than CUT3R (4.7 and 4.6~cm). The gaps narrow after GT Sim(3) alignment. These results demonstrate the Stage~I VLM's ability to predict metric-scale geometry, while the ReVSI ablations show that reconstruction pretraining benefits spatial reasoning. These evaluations support our goal of learning geometry that serves downstream reasoning. \figref{fig:recons_demo1} compares our Stage~I reconstruction with CUT3R and MapAnything. SpatialSpeak's estimate of the marked distance is closer to ground truth. Enlarged visualizations and a comparison on an additional scene are provided in~\figref{fig:recons_demo} in the appendix.

\begin{table*}[!t]
    \centering
    \caption{\textbf{Comparison with state-of-the-art methods on  ReVSI}. We evaluate SpatialSpeak and official checkpoints of SpatialStack, GeoThinker, VG-LLM, and Omni-View under the same 32-frame setting.
 Other baseline results are sourced from the ReVSI paper~\citep{zhang2026revsi}.  
}
\vspace{-0.35cm}
    \resizebox{1\linewidth}{!}{
    \begin{tabular}{r|c|ccccccc}
    & & 
    \rotatebox{30}{{Obj. Count}} &
    \rotatebox{30}{{Abs. Dist.}} &
    \rotatebox{30}{{Obj. Size}} & 
    \rotatebox{30}{{Room Size}} &
    \rotatebox{30}{{Rel. Dist.}} &
    \rotatebox{30}{{Rel. Dir.}} &
    \rotatebox{30}{{Route Plan}} \\
    \makecell[c]{{Method}} & {Avg.} & \multicolumn{4}{c}{\cellcolor{orange!10}{Numerical Answer}} & \multicolumn{3}{c}{\cellcolor{yellow!10}{Multiple-Choice Answer}} \\
    \hline

        \quad SpaceR-7B (SG-RLVR)~\citep{ouyang2025spacer} & {30.5} & {30.7}  & 34.5  & 52.0  & {18.6}  & {22.8} & 34.5  & {20.2}    \\

        \quad Spatial-MLLM-4B-135k~\citep{wu2025spatialmllm} & 40.5 & 40.7  & 45.3  & 46.8  & \makebox[1.83em][r]{--}  & {32.3} & 37.4  & \makebox[1.83em][r]{--}    \\
        \quad Spatial-MLLM-4B-820k~\citep{wu2025spatialmllm} & 40.9  & 41.5  & 40.0  & 53.1  & \makebox[1.83em][r]{--}  & {30.7}  & 39.2 & \makebox[1.83em][r]{--} \\
        \quad VST-7B-SFT~\citep{vst} & 46.4 & 35.4  & 52.6 & 67.9  & 47.2  & 49.2  & 36.9  & 35.4    \\
         VG-LLM-8B~\citep{zhenglearning} & 46.4 & 37.8 & 53.0 & 56.8 & 48.0 & 57.2 & 33.8 & 38.0 \\
         Omni-View-7B~\citep{hu2025omniview} & 48.4 & 43.1 & 52.1 & 64.2 & 52.4 & 45.9 & 38.5 & 42.2 \\
        \quad Cambrian-S-7B~\citep{yang2025cambrians} & 49.1 & 48.4  & 60.5  & 65.5 & 46.7 & 37.1 & 48.5  & 37.0   \\
        VLM-3R-7B~\citep{fan2025vlm} & 50.1 & 41.6 & 61.6 & 64.8 & 52.5 & 46.5 & 49.5 & 34.1 \\
        GeoThinker-8B~\citep{li2026thinking} & {54.0} & 37.1 & 71.5 & 66.0 & 53.0 & 61.9 & 47.5 & 41.5  \\
        SpatialStack-4B~\citep{zhang2026spatialstack} & 54.1 & 36.3 & 64.6 & 63.2 & 54.4 & 62.9 & 47.7 & 49.6 \\
        \model-4B~(Ours) & \textbf{62.8} & 64.9 & 70.2 & 62.5 & 66.0 & 71.9 & 51.1 & 52.7 \\

    \hline
    \end{tabular}
    }
\vspace{-0.5cm}
\label{tab:revsibench}
\end{table*}

\subsection{Main Results}
We report results on ReVSI and SPAR-Bench, alongside VSI-Bench under normal and scaled training settings to cover a broader range of prior methods.
\paragraph{ReVSI.}
In~\tabref{tab:revsibench}, we compare against competitive methods
on ReVSI~\citep{zhang2026revsi}, our primary spatial-reasoning benchmark.
Other methods that do not report ReVSI results are skipped.
Our method achieves an average score of 62.8, exceeding SpatialStack-4B (54.1), the highest-scoring competing method in the table, by 8.7 points.
Our method obtains the best reported result on five of the seven task categories. Compared with SpatialStack-4B, the gains include object counting (64.9 vs.\ 36.3), room size (66.0 vs.\ 54.4), and relative distance (71.9 vs.\ 62.9).
Together with the ReVSI ablations, these results support our strategy of learning multi-view geometry through reconstruction pretraining and training its use in explicit spatial reasoning through CoT with visual compensation.

\begin{table*}[!t]
    \centering
\caption{\textbf{Comparison with state-of-the-art methods on VSI-Bench}~(normal training setting). 
}
\vspace{-0.35cm}
    \resizebox{1\linewidth}{!}{
    \begin{tabular}{r|c|cccccccc}
    & & 
    \rotatebox{30}{{Obj. Count}} &
    \rotatebox{30}{{Abs. Dist.}} &
    \rotatebox{30}{{Obj. Size}} & 
    \rotatebox{30}{{Room Size}} &
    \rotatebox{30}{{Rel. Dist.}} &
    \rotatebox{30}{{Rel. Dir.}} &
    \rotatebox{30}{{Route Plan}} &
    \rotatebox{30}{{Appr. Order}} \\
    \makecell[c]{{Method}} & {Avg.} & \multicolumn{4}{c}{\cellcolor{orange!10}{Numerical Answer}} & \multicolumn{4}{c}{\cellcolor{yellow!10}{Multiple-Choice Answer}} \\
    \hline
    \rowcolor{navyblue!5}
    \hline
    
    \rowcolor{navyblue!5}
    \multicolumn{1}{l|}{\textcolor{black}{\textit{Proprietary Models (API)}}} & & & & & & & & & \\
    GPT-4o & 34.0 & 46.2 & 5.3 & 43.8 & 38.2 & 37.0 & 41.3 & 31.5 & 28.5 \\
    Gemini-1.5-Flash & 42.1 & 49.8 & 30.8 & 53.5 & {54.4} & 37.7 & 41.0 & 31.5 & 37.8 \\
    Gemini-1.5-Pro & {45.4} & {56.2} & {30.9} & {64.1} & 43.6 & {51.3} & {46.3} & {36.0} & 34.6 \\
    \hline
    \rowcolor{navyblue!5}
    \multicolumn{1}{l|}{\textcolor{black}{\textit{Open-source Models}}} & & & & & & & & & \\
    InternVL2-8B~\citep{chen2024internvl} & 34.6 & 23.1 & {28.7} & 48.2 & {39.8} & 36.7 & 30.7 & 29.9 & 39.6 \\
    InternVL2-40B~\citep{chen2024internvl} & 36.0 & 34.9 & 26.9 & 46.5 & 31.8 & 42.1 & 32.2 & 34.0 & 39.6 \\
    LongVILA-8B~\citep{chen2024longvila}  & 21.6 & 29.1 & 9.1 & 16.7 & 0.0 & 29.6 & 30.7 & 32.5 & 25.5 \\
    VILA-1.5-40B~\citep{lin2023vila}  & 31.2 & 22.4 & 24.8 & 48.7 & 22.7 & 40.5 & 25.7 & 31.5 & 32.9 \\
    LongVA-7B~\citep{zhang2024longva} & 29.2 & 38.0 & 16.6 & 38.9 & 22.2 & 33.1 & 43.3 & 25.4 & 15.7 \\
    LLaVA-Video-72B~\citep{zhang2024llava}  & 40.9 & {48.9} & 22.8 & 57.4 & 35.3 & 42.4 & 36.7 & {35.0} & {48.6} \\
    LLaVA-OneVision-72B~\citep{lillava} & 40.2 & 43.5 & 23.9 & {57.6} & 37.5 & 42.5 & 39.9 & 32.5 & 44.6 \\
    \hline
    \rowcolor{navyblue!5}
    \multicolumn{1}{l|}{\textcolor{black}{\textit{Spatial-Enhanced Models}}} & & & & & & & & & \\
    SAT-LLaVA-Video-7B~\citep{ray2025satdynamicspatialaptitude} & - & - & - & - & 47.3 & 41.1 & 37.1 & {36.1} & 40.4 \\
    SPAR-8B~\citep{zhang2025from} & 41.1 & - & - & - & - & - & - & -  & - \\
    UniUGG-3B~\citep{xu2025uniugg} & 42.2 & - & - & - & - & - & - & -  & - \\
    \quad SpaceR-7B (SG-RLVR)~\citep{ouyang2025spacer} & 45.6 & - & - & - & - & - & - & -  & - \\
    3DRS-7B~\citep{huang2025} & 45.9 & 68.7 & 34.8 & 53.6 & 56.6 & 40.9 & 43.2 & 30.4 & 39.2 \\
    SpatialLadder-3B~\citep{li2025spatialladder} & 45.7 & 63.5 & 34.3 & 61.7 & 43.9 & 45.4 & 44.8 & 35.6 & 36.4 \\
    Spatial-MLLM-4B~\citep{wu2025spatialmllm} & 47.0 & 65.3 & 34.8 & 63.1& 45.1 & 41.3 & 46.9 & 33.5 & 46.3 \\ 
    VG-LLM-4B~\citep{zhenglearning}  & 47.3 & 66.0 & 37.8 & 55.2 & 59.2 & 44.6 & 45.6 & 33.5 & 36.4 \\
        VG-LLM-8B~\citep{zhenglearning}  & {50.7} & {67.9} & {37.7} & {58.6} & {62.0} & {46.6} & 40.7 & 32.4 & {59.2} \\ 

    GeoThinker-7B~\citep{li2026thinking} & 50.5 & {69.5} & {38.5} & 57.9 & 62.2 & 45.2 & 46.2 & 31.4 & {52.6} \\
    Omni-View-7B~\citep{hu2025omniview}
    & 55.4 & 70.3 & 46.4 & 68.6 & 54.7 & 65.9 & 54.4 & 33.5 & 49.0  \\

    \model-4B~(Ours) & \textbf{63.3} & 72.2 & {56.9} & {67.6} & {62.8} & {62.5} & {86.2} & {43.3} & {55.2} \\
    \hline
    \end{tabular}
    }
\vspace{-0.2cm}
\label{tab:vsibench1}
\end{table*}

\paragraph{VSI-Bench}~(normal training setting).
Following VG-LLM~\citep{zhenglearning}, we adopt the same training sets
sampled from the LLaVA-Hound split of LLaVA-Video-178K~\citep{zhang2024llava}
and SPAR-7M~\citep{zhang2025from},
and additionally incorporate the spatial CoT data constructed as described
in~\secref{sec:stage2}.
As shown in~\tabref{tab:vsibench1}, our method achieves an average score of 63.3, the highest among the compared methods. It exceeds Omni-View-7B~(55.4) by \textbf{7.9 points}.

\begin{table*}[!t]
    \centering
\caption{\textbf{Comparison with state-of-the-art methods on VSI-Bench}~(scaled training setting). 
}
\vspace{-0.35cm}
    \resizebox{1\linewidth}{!}{
    \begin{tabular}{r|c|cccccccc}
    & & 
    \rotatebox{30}{{Obj. Count}} &
    \rotatebox{30}{{Abs. Dist.}} &
    \rotatebox{30}{{Obj. Size}} & 
    \rotatebox{30}{{Room Size}} &
    \rotatebox{30}{{Rel. Dist.}} &
    \rotatebox{30}{{Rel. Dir.}} &
    \rotatebox{30}{{Route Plan}} &
    \rotatebox{30}{{Appr. Order}} \\
    \makecell[c]{{Method}} & {Avg.} & \multicolumn{4}{c}{\cellcolor{orange!10}{Numerical Answer}} & \multicolumn{4}{c}{\cellcolor{yellow!10}{Multiple-Choice Answer}} \\
    \hline
    
    Qwen3-VL-8B~\citep{Qwen3-VL} & 59.8 & 67.5 & 52.6 & {76.2} & 62.3 & 60.6 & 52.5 & 32.5 & 73.8 \\
    VLM-3R-7B~\citep{fan2025vlm} & 60.9 & 70.2 & 49.4 & 69.2 & 67.1 & 65.4 & 80.5 & 45.4 & 40.1 \\
    Map2Thought-7B~\citep{gao2026map2thought} & 61.0 & 70.8 & 55.0 & 70.1 & 69.4 & 56.9 & 69.8 & 38.1 & 57.4 \\
    VST-7B~\citep{vst} & 61.2 &  - & - & - & - & - & - & - & - \\
    VG-LLM-8B~\citep{zhenglearning}  & {62.2} & {71.4} & {56.8} & 69.0 & {69.1} & {67.9} & 83.2 & {47.4} & {32.5} \\ 
    3DThinker-7B~\citep{chen2025think} & 63.7 &  - & - & - & - & - & - & - & - \\
    SpatialStack-4B~\citep{zhang2026spatialstack} & 67.5 & 71.0 & 55.6 & 69.1 & 68.2 & 67.3 & {84.1} & 41.2 & {83.5} \\
    Cambrian-S-7B~\citep{yang2025cambrians} & 67.5 & {73.2} & 50.5 & 74.9 & {72.2} & {71.1} & 76.2 & 41.8 & {80.1} \\
        GeoThinker-7B~\citep{li2026thinking} & {68.5} & - & - & - & - & - & - & - & - \\
    SenseNova-SI-8B~\citep{cai2025scaling} & 68.8 & 72.0 & 53.5 & 76.8 & 72.8 & 69.6 & 80.8 & 48.5 & 76.4  \\
    GeoThinker-8B~\citep{li2026thinking} & {72.6} & - & - & - & - & - & - & - & - \\
    \model-4B~(Ours)  & \textbf{73.0} & 73.8 & {60.7} & {76.2} & {79.9} & {70.7} & {88.0} & {47.9} & {86.9} \\ %
    
    \hline
    \end{tabular}
    }
\vspace{-0.5cm}
\label{tab:vsibench2}
\end{table*}

\paragraph{VSI-Bench}~(scaled training setting).
We extend the training data
to include SPAR~\citep{zhang2025from},
VLM-3R~\citep{fan2025vlm}, and VSI-590K~\citep{yang2025cambrians},
again augmented with our spatial CoT data (\secref{sec:stage2}).
As shown in~\tabref{tab:vsibench2}, SpatialSpeak achieves 73.0 on VSI-Bench with a 4B VLM backbone, compared with 72.6 for GeoThinker-8B.

\paragraph{SPAR-Bench.}
We further evaluate on SPAR-Bench~\citep{zhang2025from}, with detailed results provided in~\tabref{tab:sparbench_res_detail} of the appendix. \model{} achieves an average score of 76.0, exceeding SpatialStack-4B (72.0) and GeoThinker-8B (68.2) by 4.0 and 7.8 points, respectively. These results complement the ReVSI and VSI-Bench evaluations, supporting the effectiveness of our reconstruction-to-reasoning training strategy across a broader range of spatial tasks.

\paragraph{Qualitative Examples.}
\figref{fig:demov2} presents an example from  ReVSI~\citep{zhang2026revsi}.
Given the question ``How many blackboards are in the scene?'', SpatialSpeak enumerates two blackboard instances and expresses their estimated 3D centers, correctly predicting a count of 2, whereas the variant with neither QA-RP nor CoT-VC predicts 3.
Additional examples appear in Appendix~\secref{sec:app_quali}.

\vspace{-0.1cm}
\section{Conclusion}
\vspace{-0.2cm}
We have presented SpatialSpeak, a two-stage training framework that combines multi-view
reconstruction pretraining for local geometry and global scene context with explicit spatial reasoning supervision.
QA-RP uses local point and global object-center queries to supervise geometry through
the VLM's standard QA outputs. Spatial CoT finetuning then teaches the model to
express question-relevant geometric estimates and use them to derive answers,
with visual compensation supporting refinement when needed.
SpatialSpeak achieves state-of-the-art performance on ReVSI, VSI-Bench and SPAR-Bench.
ReVSI ablations show that local and global reconstruction supervision both contribute
to spatial reasoning, and that QA-RP increases the gains from the subsequent CoT-VC stage.
These findings support the complementary roles of learning multi-view geometry and
explicitly supervising its use in spatial reasoning within our framework.
We hope this work motivates exploration of reconstruction-augmented reasoning
in vision-language models.



\bibliography{references}
\bibliographystyle{references}
\newpage
\appendix
\section{Appendix}

\subsection{Evaluation on SPAR-Bench}
\label{sec:app_spar}

\renewcommand{\arraystretch}{1.25} %
\setlength{\tabcolsep}{3pt} %
\begin{table*}[ht]
\footnotesize
    \centering
    \small
     \caption{\textbf{Comparison with state-of-the-art models on SPAR-Bench}~\citep{zhang2025from}. Baselines include InternVL2~\citep{chen2024internvl}, InternVL2.5~\citep{chen2024expanding}, LLaVA-OV~\citep{lillava}, Qwen2-VL~\citep{Qwen2VL}, Qwen2.5-VL~\citep{Qwen2.5-VL}, LLaVA-v1.5~\citep{liu2024improved}, LLaVA-v1.6~\citep{liu2024llavanext},  Spatial-MLLM~\citep{wu2025spatialmllm}, VLM-3R~\citep{fan2025vlm}, UniUGG-3B~\citep{xu2025uniugg}, G$^2$VLM-SR~\citep{hu2025g}, GeoThinker~\citep{li2026thinking}, SenseNova-SI~\citep{cai2025scaling} and SpatialStack~\citep{zhang2026spatialstack}. Baseline results are primarily sourced from the supplementary material of G$^2$VLM~\citep{hu2025g}.}
     \vspace{-0.35cm}
    \label{tab:sparbench_res_detail}
    \resizebox{1.0\textwidth}{!}{
    \begin{tabular}{r|c|*{9}{c}|*{4}{c}|*{10}{c}}
    \specialrule{\lightrulewidth}{\abovetopsep}{0pt}
    \rowcolor{gray!8}
        Method  & \normalsize\rotatebox{75}{Avg.} & \cellcolor{orange!10}\normalsize\rotatebox{75}{Low} & 
       \cellcolor{orange!10}\scriptsize \rotatebox{75}{Depth-OC} & \cellcolor{orange!10}\scriptsize \rotatebox{75}{Depth-OC-MV} & \cellcolor{orange!10}\scriptsize \rotatebox{75}{Depth-OO} & \cellcolor{orange!10}\scriptsize \rotatebox{75}{Depth-OO-MV} & \cellcolor{orange!10}\scriptsize \rotatebox{75}{Dist-OC} & \cellcolor{orange!10}\scriptsize \rotatebox{75}{Dist-OC-MV} & \cellcolor{orange!10}\scriptsize \rotatebox{75}{Dist-OO} & \cellcolor{orange!10}\scriptsize \rotatebox{75}{Dist-OO-MV} &
       \cellcolor{yellow!10}\normalsize \rotatebox{75}{Medium} &
       \cellcolor{yellow!10}\scriptsize \rotatebox{75}{PosMatch} & \cellcolor{yellow!10}\scriptsize \rotatebox{75}{CamMotion} & \cellcolor{yellow!10}\scriptsize \rotatebox{75}{ViewChgI} &
       \cellcolor{green!5}\normalsize \rotatebox{75}{High} & \cellcolor{green!5}\scriptsize \rotatebox{75}{DistI-OO} & \cellcolor{green!5}\scriptsize \rotatebox{75}{DistI-OO-MV} & \cellcolor{green!5}\scriptsize \rotatebox{75}{ObjRel-OC-MV} & \cellcolor{green!5}\scriptsize \rotatebox{75}{ObjRel-OO} & \cellcolor{green!5}\scriptsize \rotatebox{75}{ObjRel-OO-MV} & \cellcolor{green!5}\scriptsize \rotatebox{75}{SpImag-OC} & \cellcolor{green!5}\scriptsize \rotatebox{75}{SpImag-OC-MV} & \cellcolor{green!5}\scriptsize \rotatebox{75}{SpImag-OO} & \cellcolor{green!5}\scriptsize \rotatebox{75}{SpImag-OO-MV} \\
    \specialrule{\lightrulewidth}{0pt}{\belowrulesep}

        InternVL2-2B & 28.1& 21.7& 18.1& 24.8& 23.2& 21.0& 19.5& 20.0& 26.8& 20.6& 22.8&39.7& 23.0& 5.8& 35.4&51.2& 56.0& 46.0& 31.6& 23.8& 36.0& 34.3& 17.6&22.4\\

        InternVL2-4B & 32.0 & 28.9& 23.9 & 27.2 & 20.0 & 18.1 & 42.6 & 40.2 & 31.3 & 28.2 & 29.2&49.9 & 21.0 & 16.6 & 35.7&56.8 & 55.4 & 40.3 & 36.8 & 25.2 & 28.8 & 32.3 & 21.2 & 24.7\\

        InternVL2.5-2B & 30.1 & 25.8& 39.7 & 39.7 & 12.1 & 15.0 & 30.9 & 29.6 & 20.2 & 19.0 & 22.9&37.9 & 24.3 & 6.6 & 36.4&51.5 & 56.9 & 50.3 & 33.8 & 24.1 & 27.2 & 35.2 & 26.5 & 22.4\\

        InternVL2.5-4B & 30.6& 25.7& 29.1& 33.0& 21.8& 16.8& 20.8& 26.9& 28.1& 28.8& 29.8&47.1& 33.3& 8.9& 35.2&54.1& 58.9& 35.5& 29.7& 34.6& 24.7& 31.4& 19.2&28.3\\

        InternVL2.5-8B & 36.3 & 29.5& 25.8 & 29.3 & 23.8 & 18.8 & 46.8 & 42.7 & 22.6 & 25.9 & 31.9 &61.3 & 28.0 & 6.3 &43.8 &59.7 & 56.9 & 51.8 & 44.2 & 41.6 & 36.6 & 41.6 & 22.5 & 39.5\\

        LLaVA-OV-0.5B & 29.5 & 30.1& 49.2 & 42.7 & 18.0 & 14.9 & 31.5 & 25.7 & 29.0 & 30.1 & 15.9&24.4 & 21.8 & 1.5 & 33.4&50.9 & 50.0 & 32.0 & 27.8 & 26.0 & 30.9 & 34.0 & 24.5 & 24.7\\

        LLaVA-OV-7B & 31.2 & 21.8& 30.3 & 26.9 & 18.6 & 13.9 & 10.4 & 13.6 & 31.2 & 29.3 &26.1 &38.7 & 30.3 & 9.5 &40.1 &56.5 & 55.1 & 37.3 & 48.6 & 38.2 & 30.4 & 33.7 & 26.5 & 35.0\\

        Qwen2-VL-2B & 24.6 & 19.4& 38.0 & 40.6 & 18.8 & 14.1 & 7.8 & 7.1 & 17.8 & 11.1 & 27.6&26.2 & 25.3 & 31.2 &  28.2&54.1 & 49.1 & 21.8 & 25.3 & 12.5 & 23.9 & 27.6 & 24.8 & 14.9\\

        Qwen2-VL-7B & 30.7 & 27.5& 36.0 & 35.2 & 20.8 & 12.9 & 28.7 & 30.0 & 28.2 & 28.5 & 20.4&35.4 & 20.3 & 5.7 & 37.0&59.7 & 52.4 & 30.3 & 38.5 & 41.0 & 22.0 & 28.5 & 22.5 & 38.4\\

        Qwen2.5-VL-3B  & 29.4  & 26.7  & 31.7 &34.2 & 32.1 & 17.5 & 18.4 & 22.7 & 32.1 & 24.8 & 24.9 & 39.2 & 27.3 &8.1  & 33.3 & 55.6 & 60.7 & 37.5 & 32.1& 20.2 & 21.0& 27.0 & 20.9 & 24.6\\
        
        Qwen2.5-VL-7B & 33.1 & 28.8& 31.3 & 33.7 & 22.0 & 15.0 & 42.9 & 37.7 & 23.8 & 23.6 & 23.0&33.3 & 28.8 & 6.8 & 40.3&58.2 & 51.5 & 44.8 & 50.0 & 32.1 & 33.9 & 32.9 & 27.2 & 31.9\\

        LLaVA-v1.5-7B & 23.7 & 10.9& 5.2 & 12.5 & 17.4 & 11.3 & 7.3 & 5.3 & 18.7 & 9.1 & 26.5&24.4 & 26.8 & 28.3 & 34.1&51.2 & 52.4 & 34.3 & 24.2 & 26.9 & 34.7 & 29.9 & 22.5 & 30.8\\

        LLaVA-v1.6-7B & 13.2 & 8.5& 12.1 & 0.0 & 20.4 & 0.3 & 10.8 & 0.4 & 24.3 & 0.0 & 4.8&6.6 & 7.8 & 0.0 &  20.2&51.8 & 7.7 & 6.3 & 32.1 & 6.4 & 39.5 & 10.5 & 21.5 & 5.9\\

        Spatial-MLLM-7B  & 32.2 & 29.9 & 31.9 & 22.9 & 22.8 & 16.4 & 35.9 & 38.7 & 35.5 & 34.9 & 20.3 & 34.1 & 26.8 & 0.0 & 38.1 & 54.7 & 50.9 & 39.0 & 34.6 & 24.7 & 38.7 & 41.3 & 28.8 & 30.5 \\

        VLM-3R-7B  & 43.2 & 39.8 & 47.8 & 45.6 & 40.1 & 20.6 & 42.2 & 44.3 & 40.1 & 37.5 & 28.4 & 42.0 & 30.0 & 13.3 & 51.2 & 55.9 & 59.2 & 58.8 & 53.0 & 54.6 & 47.3 & 50.6 & 30.5 & 50.7 \\
        
        SenseNova-SI-8B
        & 45.8 & - & - & - & - & -& - & - & - & - & - & - & - & - & - & - & - & - & - & - & - & - & - & - \\
        UniUGG-3B & 50.6 & 50.8 & - & - & - & -& - & - & - & - & 49.1 & - & - & - & 51.9 & - & - & - & - & - & - & - & - & - \\
        {\textsc{G$^2$VLM}\xspace}-SR-2B  & 54.9 & 60.0 & 80.3 & 73.8 & 21.4 & 18.9 & 78.4 & 75.2 & 68.4 & 63.6 & 36.3 & 27.0 & 28.2 & 53.3 & 56.5 & 53.5 & 49.1 & 76.8 & 50.0 & 68.7 & 50.5 & 52.6 & 44.4 & 63.0 \\
        GeoThinker-8B & 68.2 & - & - & - & - & -& - & - & - & - & - & - & - & - & - & - & - & - & - & - & - & - & - & - \\
        SpatialStack-4B & 72.0 & - & - & - & - & -& - & - & - & - & - & - & - & - & - & - & - & - & - & - & - & - & - & - \\
        \model-4B~(Ours) & \textbf{76.0} & 68.0 & 89.2 & 84.6 & 39.4 & 34.0 & 87.2 & 86.3 & 71.1 & 52.2 & 75.0 & 86.8 & 81.8 & 56.6 & 83.3 & 89.1 & 90.2 & 93.5 & 86.8 & 88.1 & 76.9 & 79.7 & 64.9 & 81.0 %

        \\
    \bottomrule[\lightrulewidth]
        
    \end{tabular}
}
\end{table*}

\clearpage

\subsection{More qualitative examples}
\label{sec:app_quali}
\begin{figure}[h!]
     \centering
    \begin{overpic}[width=1\columnwidth]
    {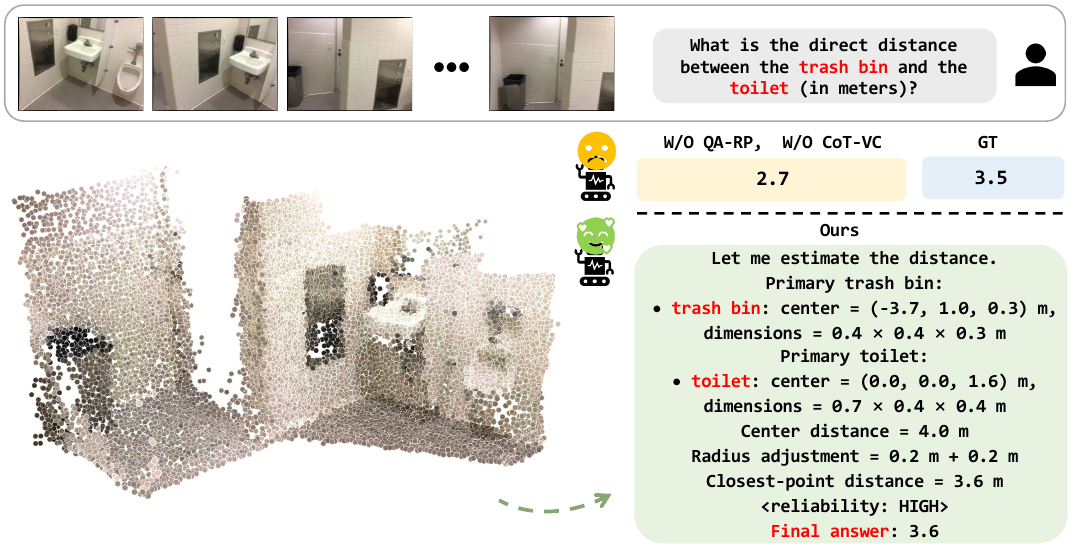}
    \end{overpic}
    \caption{
\textbf{Qualitative example}. The bottom-left panel shows the Stage~I (QA-RP) reconstruction, and the bottom-right panel shows the Stage~II (CoT-VC) reasoning output. SpatialSpeak estimates the 3D centers and dimensions of the trash bin and toilet, then approximates their closest-point distance as 3.6~m, compared with 2.7~m from the variant with neither QA-RP nor CoT-VC and a ground truth of 3.5~m. 
}
\label{fig:demo1}

\end{figure}

\begin{figure}[h!]
     \centering
    \begin{overpic}[width=1\columnwidth]
    {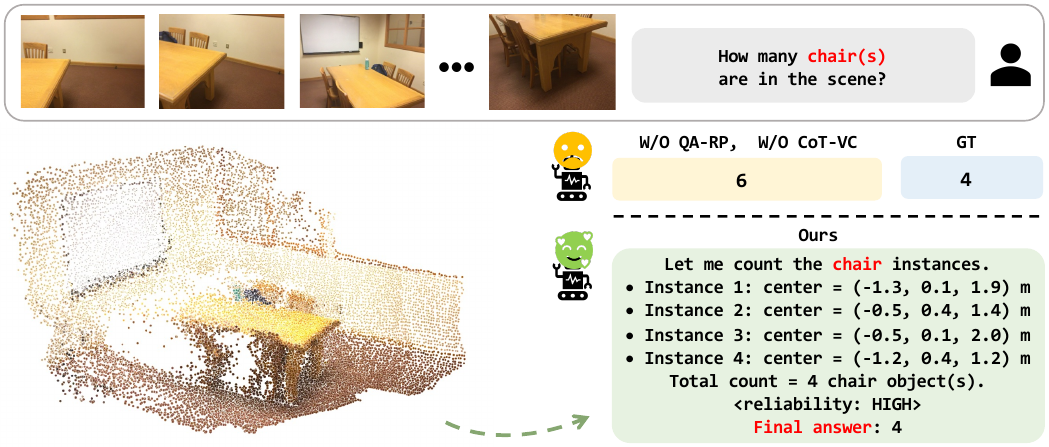}
    \end{overpic}
    \caption{
\textbf{Qualitative example}. The bottom-left panel shows the Stage~I (QA-RP) reconstruction, and the bottom-right panel shows the Stage~II (CoT-VC) response. SpatialSpeak enumerates four chair instances with distinct estimated 3D centers and predicts a count of 4, matching the ground truth, whereas the variant with neither QA-RP nor CoT-VC predicts 6.
}
\label{fig:chair_counting}

\end{figure}

\begin{figure}[h!]
     \centering
    \begin{overpic}[width=0.98\columnwidth]
    {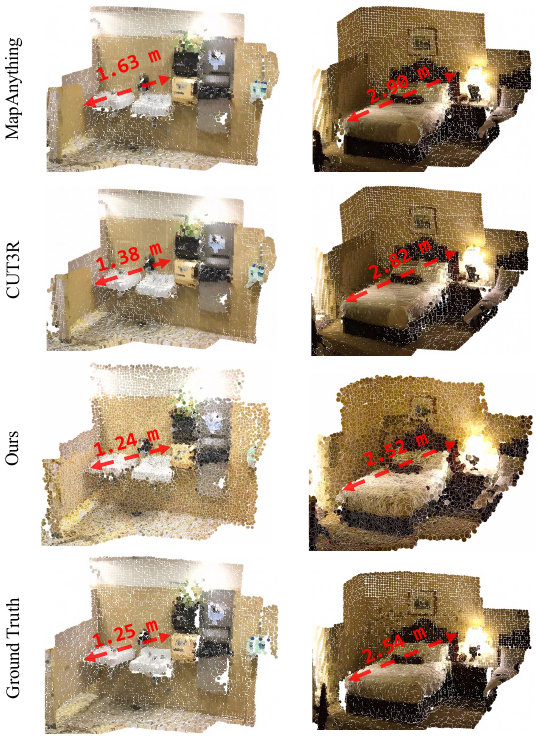}
    \end{overpic}
    \caption{
\textbf{Extended qualitative reconstruction comparisons}. Rows show MapAnything~\citep{keetha2026mapanything}, CUT3R~\citep{wang2025continuous}, SpatialSpeak after Stage~I, and ground truth from top to bottom. The left column provides enlarged visualizations of the scene in~\figref{fig:recons_demo1}, while the right column shows an additional scene. 
Red dashed arrows mark corresponding distances in meters.
SpatialSpeak's marked distances (1.24/2.52~m, left/right) are closer to the ground-truth values (1.25/2.54~m) than those of CUT3R (1.38/2.82~m) and MapAnything (1.63/2.90~m).
}
\label{fig:recons_demo}
\end{figure}

In this section, we show additional reasoning examples in~\figref{fig:demo1} and~\figref{fig:chair_counting}, together with reconstruction examples in~\figref{fig:recons_demo}. In~\figref{fig:demo1}, the model uses estimated object centers and dimensions to derive an approximate closest-point distance between a trash bin and a toilet. In~\figref{fig:chair_counting}, it enumerates four chair instances with distinct estimated 3D centers and predicts a count of 4, matching the ground truth, whereas the variant with neither QA-RP nor CoT-VC predicts 6. \figref{fig:recons_demo} extends the reconstruction comparison in~\figref{fig:recons_demo1} with enlarged visualizations of the same scene and a comparison on an additional scene. For both illustrated scenes, the distances marked in SpatialSpeak's Stage~I reconstructions are closer to ground truth than those of CUT3R and MapAnything.
\vspace{0.3cm}  

\newpage

\clearpage
\subsection{More implementation details}
\label{sec:app_imple}

In this section, we present the training hyperparameters. The configurations for Stage I and Stage II are summarized in \tabref{tab:hyper1}~and~\tabref{tab:hyper2}, respectively.

\begin{table}[h]
\centering

\caption{\textbf{Training hyperparameters in Training Stage I}.}
\vspace{-0.3cm}
\label{tab:hyper1}

\begin{tabular}{ll}
\toprule
Hyperparameter & Value \\
\midrule
accelerator model & H800 GPUs\\
accelerator count & 8 \\
global batch size & 32 \\
tune\_mm\_llm & True \\
tune\_mm\_vision & False \\
tune\_mm\_mlp & False \\
bf16 & True \\
per\_device\_train\_batch\_size & 1 \\
gradient\_accumulation\_steps & 4 \\
learning\_rate & 5e-6 \\
optim & adamw\_torch \\
num\_train\_epochs & 1 \\
warmup\_ratio & 0.03 \\
lr\_scheduler\_type & `cosine' \\
weight\_decay & 0.01 \\
\bottomrule
\end{tabular}
\vspace{0.2cm}
\end{table}

\begin{table}[h]
\centering
\caption{\textbf{Training hyperparameters in Training Stage II}.}
\vspace{-0.3cm}
\label{tab:hyper2}
\begin{tabular}{ll}
\toprule
Hyperparameter & Value \\
\midrule
accelerator model & H800 GPUs \\
accelerator count & 8 \\
global batch size & 64 \\
tune\_mm\_llm & True \\
tune\_mm\_vision & False \\
tune\_mm\_mlp & False \\
bf16 & True \\
per\_device\_train\_batch\_size & 1 \\
gradient\_accumulation\_steps & 8 \\
learning\_rate & 1e-5 \\
optim & adamw\_torch \\
num\_train\_epochs & 1 \\
warmup\_ratio & 0.03 \\
lr\_scheduler\_type & `cosine' \\
weight\_decay & 0.01 \\
\bottomrule
\end{tabular}
\vspace{0.2cm}
\end{table}

\end{document}